\documentclass[journal]{IEEEtran}
\usepackage{cite}
\usepackage{amsmath,amssymb,amsfonts}
\usepackage{graphicx}
\usepackage{textcomp}
\usepackage{xcolor}
\usepackage{hyperref}
\usepackage{algorithm}
\usepackage{caption}
\usepackage{algpseudocode}
\usepackage{listings}
\usepackage{xcolor}
\usepackage{array}
\usepackage{adjustbox}
\usepackage{multirow}
\usepackage{threeparttable}
\usepackage{tabularx}
\usepackage{booktabs} 
\def\BibTeX{{\rm B\kern-.05em{\sc i\kern-.025em b}\kern-.08em
    T\kern-.1667em\lower.7ex\hbox{E}\kern-.125emX}}
\definecolor{codegreen}{rgb}{0,0.6,0}
\definecolor{codegray}{rgb}{0.5,0.5,0.5}
\definecolor{codepurple}{rgb}{0.58,0,0.82}
\definecolor{backcolour}{rgb}{0.95,0.95,0.92}

\lstdefinestyle{mystyle}{
    backgroundcolor=\color{backcolour},   
    commentstyle=\color{codegreen},
    keywordstyle=\color{magenta},
    numberstyle=\tiny\color{codegray},
    stringstyle=\color{codepurple},
    basicstyle=\ttfamily\footnotesize,
    breakatwhitespace=false,         
    breaklines=true,                 
    captionpos=b,                    
    keepspaces=true,                 
    numbers=left,                    
    numbersep=5pt,                  
    showspaces=false,                
    showstringspaces=false,
    showtabs=false,                  
    tabsize=2
}

\begin{document}

\title{DragonVerseQA: Open-Domain Long-Form Context-Aware Question-Answering}

\author{Aritra Kumar Lahiri,~\IEEEmembership{Toronto Metropolitan University, Toronto, CA, aritra.lahiri@torontomu.ca}\\
      Qinmin Vivian Hu,~\IEEEmembership{Toronto Metropolitan University, Toronto, CA, vivian@torontomu.ca }}


\maketitle

\begin{abstract}
This paper proposes a novel approach to develop an open-domain and long-form Over-The-Top (OTT) Question-Answering (QA) dataset, DragonVerseQA, specifically oriented to the fantasy universe of "House of the Dragon" and "Game Of Thrones" TV series. Most existing QA datasets focus on short, fact-based answers sourced almost solely from Wikipedia articles, devoid of depth and contextual richness for sophisticated narrative understanding. We curate a dataset that combines full episode summaries sourced from HBO and fandom wiki websites, user reviews from sources like IMDb and Rotten Tomatoes, and high-quality, open-domain, legally admissible sources, and structured data from repositories like WikiData into one dataset. The dataset provides a multi-dimensional context, reflecting complex character dynamics and plot developments from these varied sources. That means, on equal footing, only after heavy data preprocessing and filtering methods will meaningful, non-spam unbiased reviews be available in this enriched dataset. The comprehensive insights are given through the long-form answers generated from this enriched context. This is what makes this valuable dataset for improving conversational AI, narrative analysis, sentiment analysis, summarization techniques, and relation extraction.

A comparative analysis with state-of-the-art QA datasets such as SQuAD 2.0, TriviaQA, and Natural Questions brings to light the unique advantages of our dataset in terms of contextual complexity and answer length. Detailed reviews add layers to audience sentiment and narrative interpretation, raising the bar for domain-specific QA with a new quality benchmark. Our work also allows a deeper understanding of entertainment-industry content and opens the door to more knowledgeable and creative AI-driven interactions within digital media environments.

\end{abstract}


\section{Introduction}
The rapid evolution in Natural Language Processing (NLP) has fueled the development of advanced question-answering (QA) systems designed to understand and respond to human inquiries accurately. The substantial foundations set by existing benchmark datasets, notably the \textit{Stanford Question Answering Dataset}(SQuAD)\cite{b4}, Google's \textit{Natural Questions}(NQ)\cite{b6}, and \textit{TriviaQA}\cite{b5} allowed for pretty significant advancements in this domain. These datasets are, however, majorly short, fact-based answers derived majorly from Wikipedia articles. Hence, their applications in more nuanced and contextually rich narratives, such as those from the entertainment industry, remain limited.

Particularly, Long-form QA datasets like ELI5\textit{(Explain Like I'm 5)}\cite{b3} are sourced from Reddit's community for the same name and provide expansive explanations on various general knowledge topics. ELI5 captures the general spirit of user-generated content and gives very verbose answers that should make some sense to a broader audience. However, the quality and contextual coherence are pretty varied in ELI5 because these are colossal user contributions. This again brings the need for domain-specific datasets that afford high-quality contextually aware responses, especially in more complex narratives for media and entertainment.

\begin{figure}
\centering
\includegraphics[width=0.45\textwidth]{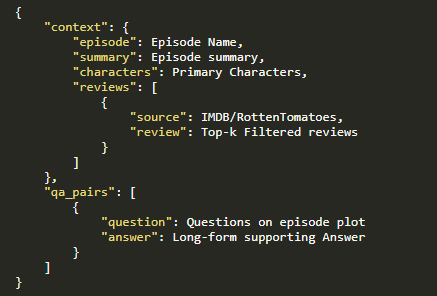}
    \vspace{-1mm}\caption{DragonVerseQA - Dataset structure}
    \label{figure: OTTQA Structure}
\end{figure}

In its main series, \textit{"Game of Thrones"} and its prequel \textit{"House of the Dragon"} by HBO, a tapestry of intricate characters, plots, and relationships warrants further comprehension and more in-depth contextual understanding. State-of-the-art long-form QA datasets struggle to capture the complexity of narratives like those found in Game of Thrones and House of the Dragon, as they typically focus on atomic, categorical facts rather than the intricate storylines crucial to such series. This paper introduces a novel QA dataset, \textbf{DragonVerseQA}, specifically designed for the fantasy universe genre in OTT content. DragonVerseQA integrates data from diverse sources, including episode summaries and guides from official websites like HBO and Fandom wikis, user reviews from platforms like IMDb and Rotten Tomatoes, and structured data from repositories like WikiData. The dataset circumvents copyright and infringement concerns by ensuring the use of legally permissible data. As shown in Fig. \ref{figure: OTTQA Structure}, the dataset's format pairs multi-dimensional contexts with corresponding QA pairs. The context includes each episode title from House of the Dragon Season 1 and Game of Thrones Season 8, along with its concise episode summary. Additionally, for each episode, the top-k (k = 10 in our implementation) high-quality reviews are curated from IMDb and Rotten Tomatoes to enrich the context and support more robust QA generation.

The primary research objective was to create a comprehensive, context-rich QA dataset that helps bridge the gap between traditional QA datasets and the requirements characterizing complex narrative understanding. This new dataset will be based on minute episode summaries, structured character information, and curated audience reviews, ensuring QA systems with high potential to answer long-form questions relevant to the context.

\subsection{Research Contribution}
Overall, our research contributions in this paper can be highlighted as follows:
\begin{itemize}
    \item \textit{Rich, Multi-Dimensional Context} - TV series like \textit{Game of Thrones} and \textit{House of the Dragon} feature complex, evolving plots and character dynamics that require a deep understanding of context. Integrating multi-dimensional context uniquely enables handling these narrative intricacies within the dataset.
    \item \textit{Long-Form Answers with Narrative Depth} - We introduce long-form answers, capturing the nuances of narrative content. High-quality QA pairs allow users to grasp the storyline of each episode in context, offering a seamless experience by following along with the dataset.
    \item \textit{Context-Aware Question Answering} - Our dataset supports generating context-aware QA pairs by predicting answer spans first, followed by generating questions based on the episode summaries, ensuring that the questions are tightly coupled with the narrative context.
    \item \textit{Domain-specific Insights with Supportive Knowledge Graph} - Harboring an entertainment-domain focus, this dataset fills a gap between the more general QA datasets and media/narrative studies. We also visualize the relationship between the episodes, characters, and user reviews with a knowledge graph to comprehensively comprehend this OTT-based TV series domain.
\end{itemize}

\subsection{Potential Applications}
This dataset can be an essential source for numerous areas within NLP research. The potential research applications are summarized below - 

1. \textit{Enhancing Conversational AI}: It can improve chatbots and AI assistants designed to discuss TV shows by providing richer contextual knowledge, including character relationships and intricate plot details.
   
2. \textit{Narrative Understanding}: This dataset enables the development of models that can analyze and comprehend complex narratives and character dynamics, which is crucial for AI applications focused on story-based content.

3. \textit{Sentiment Analysis}: The inclusion of user reviews supports research in sentiment analysis, helping to study audience reactions and emotional responses to specific episodes, characters, and storylines.

4. \textit{Relation Extraction and Knowledge Graphs}: It can contribute to constructing more sophisticated knowledge graphs that map the intricate relationships between characters and events, improving query systems for entertainment databases.

In summary, this research sets a new benchmark for OTT content by introducing the first QA dataset to capture the intricate complexities of TV series narratives. We believe this work will push QA systems towards better interpretation and engagement with complex media content, paving the way for more advanced and immersive AI-driven applications in digital media environments.

\section{Related Works}

Early works that have significantly contributed to the handling and processing of domain-specific datasets include the MovieQA dataset\cite{b7} and the TVQA dataset\cite{b8}. These datasets advanced the development of models capable of understanding and answering questions related to plot, context, and narrative nuances in movies and TV series. Research on Over-The-Top (OTT) media content, such as television series streamed over the internet, has primarily focused on recommendation systems and viewership pattern analysis. Notable studies include the work by Gomez-Uribe and Hunt\cite{b9} at Netflix on personalized video content recommendations, which indirectly impacts how viewers interact with OTT content via question-answering and information retrieval systems. Ramakrishna et al.\cite{b10} focused on script and subtitle analysis, mainly extracting structured data from unstructured multimedia content, a technique crucial for developing datasets that require a deep understanding of dialogues and narrative structures. Research by Bender et al. \cite{b11} emphasizes the importance of cultural and contextual awareness in AI systems, underscoring the need for these dimensions when designing QA systems for narratives from specific cultural or thematic contexts, such as "House of the Dragon" and "Game of Thrones."

When comparing with prominent open-domain long-form QA datasets, several factors come into play, such as context dependency, source material, question types, and answer formats. Datasets like ELI5 have driven recent advances in long-form question answering (Explain Like I’m Five)\cite{b3}, which focuses on generating explanatory answers to open-domain questions. ELI5, built from user-generated Reddit questions, spans various topics, contrasting with the structured script and episode summaries in DragonVerseQA. While ELI5 requires broad, educational explanations, often necessitating comprehensive background information, DragonVerseQA focuses on detailed narrative and plot elements within a fictional universe. Both datasets encourage long-form answers, but DragonVerseQA emphasizes narrative coherence and storytelling.

NarrativeQA\cite{b12} is another notable dataset that evaluates story comprehension, requiring answers based on summarized stories from books and movie scripts. While NarrativeQA and House of the Dragon QA focus on understanding narratives, the latter is closely tied to ongoing plotlines and character development within serialized TV shows. On the other hand, Natural Questions (NQ)\cite{b6}, created by Google researchers, presents real-user queries from Google Search and requires models to answer based on Wikipedia pages. Unlike DragonVerseQA, NQ includes short and long answers, necessitating extensive content interpretation.

Similarly, MS MARCO (Microsoft Machine Reading Comprehension)\cite{b13} addresses real-world information needs using anonymized user queries from Bing’s search engine. Still, its answers are primarily short and fact-based, whereas DragonVerseQA emphasizes deeper narrative exploration. SQuAD 2.0\cite{b4} is another reading comprehension dataset with crowd-sourced questions posed on Wikipedia articles, but it focuses on factual extraction, making it less suitable for interpreting complex fictional storylines. HotpotQA\cite{b14} requires synthesizing information across multiple documents, which aligns with DragonVerseQA’s demand for understanding interconnected plot points. However, the latter focuses on narratives within the same series.

Conversational QA datasets like QuAC\cite{b15} and CoQA\cite{b16} emphasize dialogue-based context retention, differing from DragonVerseQA, which provides rich narrative contexts for TV series episodes. Similarly, TriviaQA\cite{b5} Long centers on trivia questions with evidence from Wikipedia and web searches but lacks the narrative depth found in episodic content.

DragonVerseQA introduces new challenges in understanding long-form narratives, character development, and thematic complexity. Compared to more general QA datasets, it is explicitly designed for serialized TV narratives, offering valuable insights for developing QA models that handle intricate storytelling structures within fictional worlds. Its detailed focus on plot exposition and episode-specific analysis sets a new standard for QA systems in the media domain.

\section{Methodology}
This section is divided into detailed phases covering specific tasks and technical implementations.

\subsection{Data Collection}
 Our data compilation focuses on public domain texts and legally permissible sources to gather episode summaries, guides, user reviews, and character lists. The data sources are categorized as follows:
\begin{itemize}
    \item \textit{Episode Summaries and Guides} - Collected from official network websites (HBO) and fan websites, adhering to robots.txt and terms of service.
    \item \textit{Reviews} - Extracted from authorized sources like IMDb and Rotten Tomatoes, which are often publicly accessible and don't infringe on copyright.
    \item \textit{Structured Data} - Collated from databases like WikiData.
\end{itemize}

During data gathering, we read the robots.txt for adhering to the web scraping rules and regulations and fetch the contents of the URL using BeautifulSoup\cite{b18} library accordingly if scraping is allowed as denoted in Algorithm \ref{alg:webscraper}. 

\begin{algorithm}
\caption{Fetch URL Content}
\label{alg:webscraper}
\begin{enumerate}
  \item \textbf{Input:} URL
  \item Initialize RobotFileParser
  \item Set URL for \texttt{robots.txt}
  \item Read \texttt{robots.txt}
  \item \textbf{if} scraping not allowed \textbf{then}
  \begin{enumerate}
    \item Raise exception "Scraping not allowed by robots.txt"
  \end{enumerate}
  \item \textbf{endif}
  \item Fetch URL content
  \item Parse content as HTML
  \item \textbf{return} Parsed HTML content
\end{enumerate}
\end{algorithm}

Algorithm \ref{alg:episode_summaries} provides a sample pseudocode illustration for collating the episode summaries from HBO's official website\footnote{https://www.hbo.com/house-of-the-dragon/episodes}, \footnote{https://www.hbo.com/game-of-thrones/season-8}. We follow the same procedure for extracting user reviews shown in Algorithm \ref{alg:episode_summaries}. Additionally, we gather structured data for character lists, timelines, and storyline progression from WikiData. Some example repositories for character lists are obtained from this JSON data \footnote{https://www.wikidata.org/wiki/Special:EntityData/Q72930269.json} and storyline progression from this JSON \footnote{https://www.wikidata.org/wiki/Special:EntityData/Q23572.json}. This helps create more informative and high-quality context-aware QA pair generation.

\begin{algorithm}
\caption{Collect Episode Summaries}
\label{alg:episode_summaries}
\begin{enumerate}
  \item \textbf{Input:} base\_url
  \item html\_content $\gets$ Fetch URL Content(base\_url)
  \item summaries $\gets$ empty list
  \item \textbf{for} each episode in html\_content \textbf{do}
  \begin{enumerate}
    \item title $\gets$ extract title from episode
    \item summary $\gets$ extract summary from episode
    \item Append \{title, summary\} to summaries
  \end{enumerate}
  \item \textbf{end for}
  \item \textbf{return} summaries
\end{enumerate}
\end{algorithm}

\subsection{Data Pre-Processing}
We clean and normalize the data for the data preprocessing step to ensure consistency and usability. This involves removing extraneous content like references and normalizing whitespaces. We remove any Personally Identifiable Information (PII) for structured data and user reviews to ensure anonymization and privacy. We also standardize the format to ensure all entries have a consistent structure.

\subsection{Spam Detection and Bias Filtering in User Reviews}

To identify spam reviews, we collect the raw review data using Algorithm 1 from IMDb and Rotten Tomatoes and perform data preprocessing. The steps include removing HTML tags, URLs, emojis, and special characters from review comments. After that, tokenization and lemmatization are carried out to reduce the words to their base or root forms. 

For the initial labeling of the review data, we use a pre-trained open-source BERT-based spam detection model from Huggingface\cite{b19} to prepare the training set. Similarly, we apply a BERT-based sentence classification model for sentiment analysis and detecting potential bias in the reviews. Once the base model is trained, we use this model to predict labels for the remaining unlabeled data of the user reviews by iterating the process to refine the model continually. Algorithm \ref{alg:spam detection} describes the semi-supervised learning approach for the spam detection flow. The steps involved splitting the initial labeled data into feature vectors. After that, we train the initial spam detection model using "RandomForestClassifier" and evaluate the metrics. Then, we use the trained model to predict pseudo-labels and select confident predictions based on probability thresholds. Finally, the pseudo-labeled data is merged with the original labeled dataset to form a combined dataset of reviews.

\begin{figure}
\centering
\includegraphics[width=0.45\textwidth]{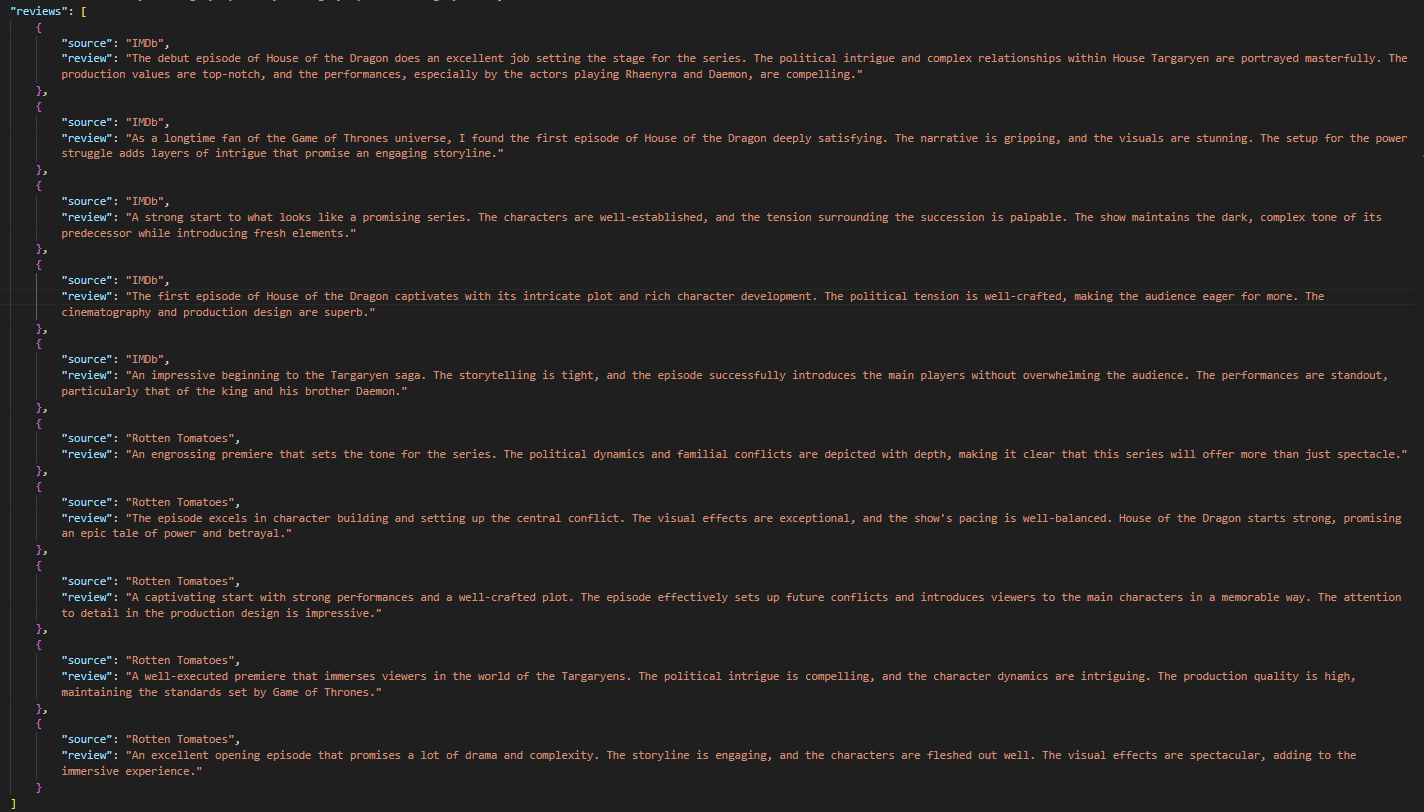}
    \vspace{-2mm}\caption{Filtered User Reviews}
    \label{figure:userreviews}
\end{figure}

We follow a similar pattern of a semi-supervised learning approach for bias filtering to eliminate overly biased reviews. We train an SVM(Support Vector Machine) model using sentiment features to classify whether a review is biased. Then, we also calculate the sentiment polarity score using TextBlob library\cite{b20} and filter out polarities between -0.5 and 0.5 to avoid extreme reviews. This ensures the quality of the final user reviews merged into our dataset. Some of the filtered user reviews collected for Episode 1 of the "House of The Dragon" TV series are shown in Figure. \ref{figure:userreviews}.

\begin{algorithm}
\caption{Semi-Supervised Learning for Spam Detection}
\label{alg:spam detection}
\textbf{Input:} Initial labeled data \(\mathcal{L}\), Unlabeled data \(\mathcal{U}\) \\
\textbf{Output:} Combined labeled data \(\mathcal{C}\) \\
\textbf{Procedure:}
\begin{enumerate}
    \item Split initial labeled data:
        \begin{itemize}
            \item \(X_{\text{train}}, X_{\text{test}}, y_{\text{train\_spam}}, y_{\text{test\_spam}}\)
        \end{itemize}
    \item Convert to feature vectors:
        \begin{itemize}
            \item \(\text{vectorizer} \gets \text{TfidfVectorizer()}\)
            \item \(X_{\text{train\_vec}} \gets \text{vectorizer.fit\_transform}(X_{\text{train}})\)
            \item \(X_{\text{test\_vec}} \gets \text{vectorizer.transform}(X_{\text{test}})\)
        \end{itemize}
    \item Train initial spam detection model:
        \begin{itemize}
            \item \(\text{spam\_model} \gets \text{RandomForestClassifier()}\)
            \item \(\text{spam\_model.fit}(X_{\text{train\_vec}}, y_{\text{train\_spam}})\)
        \end{itemize}
    \item Predict and evaluate:
        \begin{itemize}
            \item \(y_{\text{pred\_spam}} \gets \text{spam\_model.predict}(X_{\text{test\_vec}})\)
        \end{itemize}
    \item Predict labels for unlabeled data:
        \begin{itemize}
            \item \(X_{\text{unlabeled\_vec}} \gets \text{vectorizer.transform}(\mathcal{U})\)
            \item \(\hat{Y}_\mathcal{U} \gets \text{spam\_model.predict}(X_{\text{unlabeled\_vec}})\)
        \end{itemize}
    \item Add confident pseudo-labels:
        \begin{itemize}
            \item \(\hat{P}_\mathcal{U} \gets \text{spam\_model.predict\_proba}(X_{\text{unlabeled\_vec}})\)
            \item \(\text{confident\_indices} \gets [i \mid i, p \text{ in enumerate}(\hat{P}_\mathcal{U}[:, 1]) \text{ if } p > 0.9 \text{ or } p < 0.1]\)
            \item \(\mathcal{L}_\text{pseudo} \gets \{(\mathcal{U}[i], \hat{Y}_\mathcal{U}[i]) \mid i \in \text{confident\_indices}\}\)
        \end{itemize}
    \item Combine with initial training data:
        \begin{itemize}
            \item \(\mathcal{C} \gets \mathcal{L} \cup \mathcal{L}_\text{pseudo}\)
        \end{itemize}
\end{enumerate}

\end{algorithm}
\subsection{Contextual Data Integration}
\begin{figure}
\centering
\includegraphics[width=0.45\textwidth]{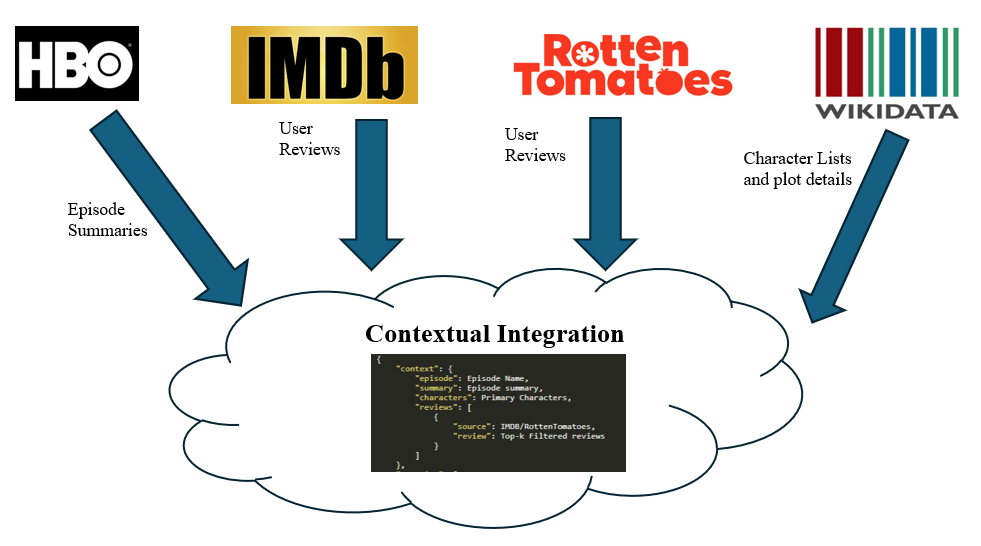}
    \vspace{-2mm}\caption{Contextual Data Integration}
    \label{figure:contextual_integration}
\end{figure}

In this step, the data collected from multiple sources is collated together to form a cohesive, unified, and structured format that can be summarized for context-aware question-answer pair generation. Figure \ref{figure:contextual_integration} refers to integrated data format from multi-dimensional resources.

To prepare this context for question-answer pair generation, we must ensure that each context chunk is meaningful and contextually rich. We apply zero-shot abstractive summarization with GPT-3 to create contextually aware summaries.

Given each text chunk \(k_{i,j} \in K_{C_j}\), the summarization pipeline leverages GPT's completion API to generate zero-shot summarization of the chunked data:

\begin{equation}
sum_{i,j} = \text{{GPT3\_Summarization}}(k_{i,j})
\end{equation}

This accommodates potential exceedance in GPT’s maximum token count by subdividing chunks based on semantic sentence similarity \(sim_{i,j}\), identifying split points (\(sp\)) to produce two sub-chunks. 

\begin{equation}
Sum_j = \text{{GPT3\_Summarization}}(sum\_{1,j} + sum\_{2,j} + \cdots)
\end{equation}


This methodology yields an abstract summary for each context chunk. The procedure is uniformly applied across the data collection \(D\), ensuring each context chunk is summarily represented. Ultimately, these summaries are amalgamated into a comprehensive summary using GPT's API, denoted as:

\[
\text{Sum} = \text{{GPT3\_Summarization}}(\text{Sum}_1 + \text{Sum}_2 + \cdots)
\]

This highly scalable approach ensures a cohesive structure by categorizing and labeling text data for enhanced narrative construction.

\subsection{Context-Aware Question Answering}

\begin{figure}
\centering
\includegraphics[width=\columnwidth]{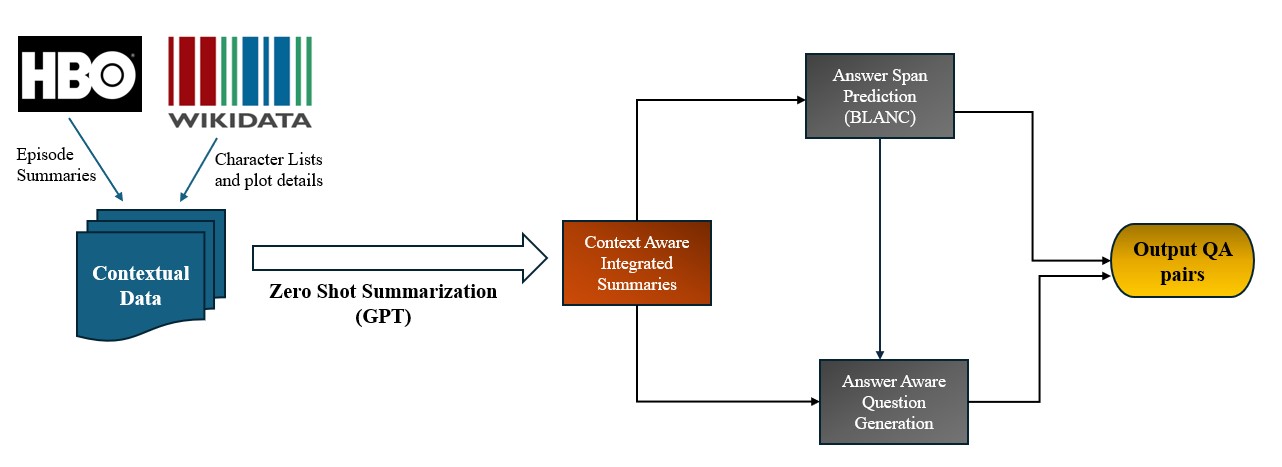}
    \vspace{-2mm}\caption{Context-Aware Question Answering: Architecture}
    \label{figure: Context-Aware QA System Diagram}
\end{figure}

Once the context-aware summaries are generated, the next step involves generating question-answer pairs. Our method for generating the question-answer pairs can be summarized broadly into three steps - (1) answer span prediction using BLANC\cite{b21}, (2) question generation based on predicted answer spans, and (3) output of the final combined QA pairs. 
The system architecture of our context-aware question-answering mechanism is shown in Figure \ref{figure: Context-Aware QA System Diagram}.

\subsubsection{\textbf{Answer Span Prediction}}

For context-aware question-answering, we use the Block Attention model \cite{b21} for predicting answer spans. This model is based on two key principles: 1) determining the probability that a word in a sentence is part of the context, represented by a hidden state, and 2) ensuring that this probability captures spatial proximity between adjacent words. The probability of an answer span within a given context is denoted by \( p_{\text{soft}}(w_i \in C) \), which serves as a soft label for the auxiliary context prediction task. Mathematically, this is expressed as denoted in Equation \ref{eq:blanc} -
\begin{equation}
p_{\text{soft}}(w_i \in C) = 
\begin{cases} 
1.0 & \text{if } i \in [s_a, e_a] \\
q^{|i-s_a|} & \text{if } i < s_a \\
q^{|i-e_a|} & \text{if } i > e_a
\end{cases}
\label{eq:blanc}
\end{equation}

In this equation, q is a hyperparameter between 0 and 1, controlling the decay rate based on the distance from the answer span. To maintain computational efficiency, the model applies this formula only to words within a specified window size around the answer span. Words outside this window are assigned a soft probability \( p_{\text{soft}}(w_i \in C) \) of 0, thereby excluding them from the context calculation.

\subsubsection{\textbf{Question Generation}}

For the Question Generation part, we adopt our Answer-Aware Question Generation (AAQG) model, applying the principle of sequential question generation, imbibing the BERT-HLSQG \cite{b2} model. For the initial fine-tuning, we train using the ELI5 dataset, designed for long-form question-answering.

Given the context paragraph \textbf{input} $\mathbf{C} = [c_{1}, ..., c_{|I|}]$ and the selected answer phrase $\mathbf{A} = [a_{1}, ..., a_{|A|}]$,
we have the question-answer pairs as our \textbf{output} to be represented as:
\begin{equation} \label{eq:qa}
QA ={[A:A_{1}….An, Q: Q_{1}….Q_{n}]}
\end{equation}

Then, we formulate a new context $C^{\prime}$, with \textbf{Highlight Tokens [HL]} as the input sequence to the AAQG model.
\begin{equation}
 C^{\prime} =  [c_{1}, c_{2}, ..., [HL], a_{1}, ..., a_{|A|} , [HL], ..., c_{|C|} ]
\label{eq:AAQG}
\end{equation}

Given the above $C^{\prime}$, the input sequence S to the AAQG model is denoted by 
\begin{equation}
 {S_{i} = ([CLS];I_{[0]}; [SEP]; q_{1}, …., q_{i}; [MASK])}
\end{equation}

In order to generate ${Q_{i}}$ in Equation \ref{eq:qa}, we apply AAQG to calculate the label probabilities as:
\begin{equation}
 Q_{i}=argmax_{w}P_{r}(w|S_{i})
\end{equation}
where $P_{r}(w|S_{i}) \in softmax(h_{[MASK]} W_{AAQG} + b_{AAQG})$. Note that we take the final hidden state for the last token \textbf{[MASK]} in the input sequence and connect it to the next connected layer $\mathbf{{W_{AAQG}}}$.

The reason for inserting a new token [HL] in Equation \ref{eq:AAQG} is based on the requirement to correctly interpret the answer span in case of longer summary chunks. This removes possible ambiguity and improves the accuracy and quality of the generated QA pairs. 

\begin{figure}
\centering
\includegraphics[width=\columnwidth]{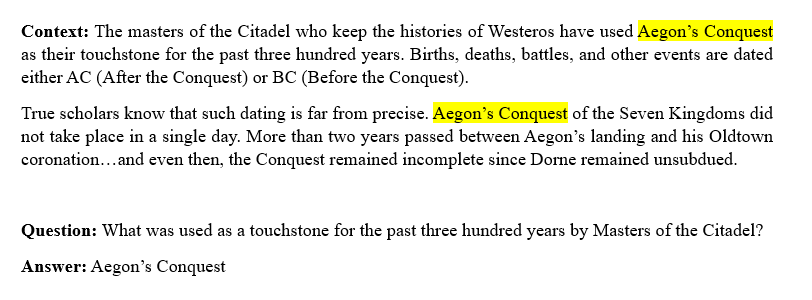}
    \vspace{-2mm}\caption{Ambiguity Resolution during Question Generation}
    \label{figure:hlsqg}
\end{figure}

However, another issue arises in duplicate answer spans within the same context. To illustrate, we look into the example portrayed in Figure \ref{figure:hlsqg}, where we observe multiple answer spans in a passage that can match the answer text for the generated question. However, based on the paragraph text, the first occurrence of the word "Aegon's Conquest" matches the context of the question only, and the second one is irrelevant. Hence, it is essential to understand the context relevant to the given question to avoid this discrepancy. This is addressed using the block attention method, BLANC (BLock AttentioN for Context prediction) \cite{b21} described in the previous sub-section, that explicitly forces the QA model to predict the context. The context prediction task predicts soft labels generated from given answer spans. The block attention mechanism computes the probability of each word in the summarized chunk for inclusion consideration in the context with negligible extra parameters and inference time.

\subsubsection{\textbf{Question Answer Pair Output}}
The final part of the implementation involves generating the QA pairs using the task-based prefixes. Three pipeline tasks generate the QA pairs - i) \textbf{single-qa}: It generates only a single pair of qa. ii) \textbf{multi-qg}: This option is answer-agnostic and generates multiple questions from the context summary. iii) \textbf{e2e-qa}: This generates all the possible long-form QA pairs from the context summary.

\begin{figure}
\centering
\includegraphics[width=\columnwidth]{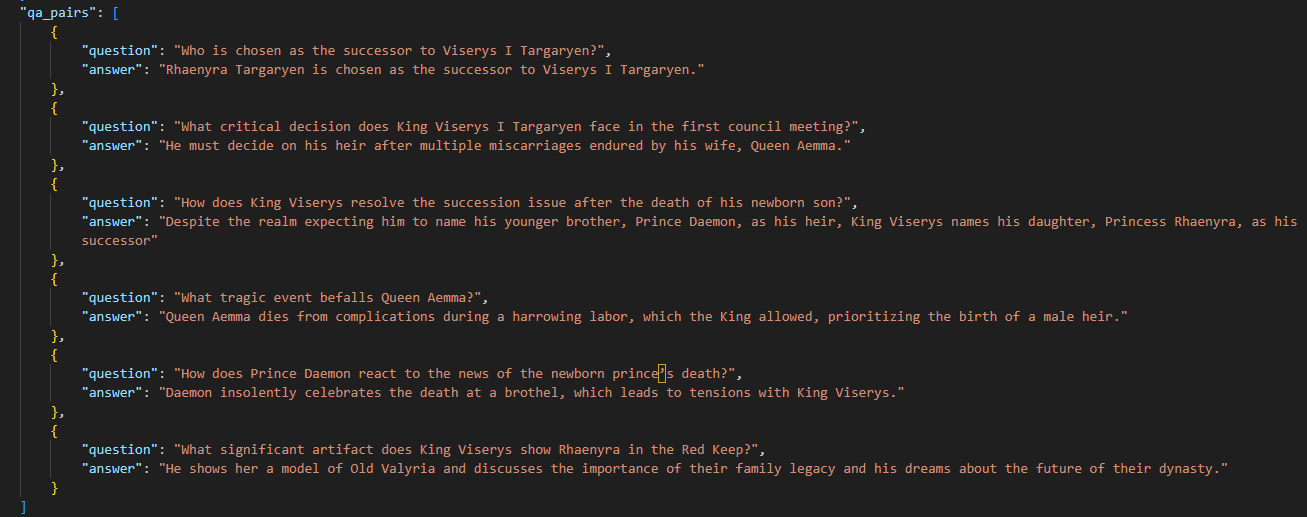}
    \vspace{-2mm}\caption{DragonVerseQA Sample QA pairs}
    \label{figure:qaepisode1}
\end{figure}

In our dataset generation part, we use \textbf{e2e-qa} task prefix to obtain the final output QA pair for each episode to integrate with the multidimensional dataset structure. Figure \ref{figure:qaepisode1} is a sample illustration of the generated long-form QA pairs from Episode 1 of the "House of The Dragon" TV series.

We also incorporate Spam Detection and Bias Filtering in our pipeline for each QA pair before appending it to the dataset. Algorithm 4 summarizes the steps of the whole process. As described in Section III. C. after pre-processing the QA pair, we pass it through the spam detection model and implement the spam classification. The non-spam QA pairs are then further analyzed for bias filtering, and finally, the cleaned, unbiased QA pairs are added to the DragonVerseQA dataset.

\begin{algorithm}
\caption{Detect Spam and Bias in QA Pair}
\begin{algorithmic}[1]
\label{alg:biasspamfilterqa}
\Function{DetectSpam}{$\text{text}$}
    \State $\text{text\_vector} \gets \text{vectorizer.transform}([\text{text}])$
    \State \Return $\text{model.predict}(\text{text\_vector})[0]$ \Comment { 0 or 1}
\EndFunction

\Function{DetectBias}{$\text{text}$}
    \State $\text{sentiment} \gets \text{extract\_sentiment}(\text{text})$
    \State $\text{sent\_vector} \gets [[\text{sentiment}]]$
    \State \Return $\text{bias\_model.predict}(\text{sent\_vector})[0]$ \Comment{0 or 1}
\EndFunction

\Function{ProcessQAPair}{$\text{QA\_Pairs}$}
    \For{each $QA\_pair$ \textbf{in} $QA\_Pairs$}
        \State \textbf{Pre-Processing:}
        \State \hspace{\algorithmicindent} $Cleaned\_QA \gets \text{Clean}(QA\_pair)$
        
        \State \textbf{Spam Detection:}
        \State \hspace{\algorithmicindent} $SpamStatus \gets \text{DetectSpam}(Cleaned\_QA)$
        \If{$SpamStatus = \text{TRUE}$}
            \State Flag $QA\_pair$ as spam \textbf{or} discard
            \State \textbf{continue}
        \EndIf
        
        \State \textbf{Bias Filtering:}
        \State \hspace{\algorithmicindent} $BiasStatus \gets \text{DetectBias}(Cleaned\_QA)$
        \If{$BiasStatus = \text{TRUE}$}
            \State Flag $QA\_pair$ for review or correction
        \EndIf

        \State \textbf{Post-Processing and Output:}
        \State \hspace{\algorithmicindent} Output $Cleaned\_QA$
    \EndFor
\EndFunction
\end{algorithmic}
\end{algorithm}



\section{Experimental Results}
\subsection{DragonVerseQA dataset}
Our current version generated 3200 high-quality long-form QA pairs covering all ten episodes of Season 1 of the \textit{House of the Dragon} and six episodes of Season 8 of the \textit{Game of the Thrones} TV series. The dataset statistics by each episode are denoted in Table \ref{table:qadataset1} and Table II. Table \ref{tab:summary_metrics} represents the summarized QA metrics of the overall dataset. We represent the overarching view of the narrative focus and the analytical breakdown of the content of the QA pairs in terms of character prominence and thematic consistency, concentrating on context awareness. Along with these, we have included ten filtered high-quality user reviews extracted each from IMDb and RottenTomatoes, which add up to 200 reviews for the overall dataset. The prototype version of the DragonVerseQA dataset has been released\footnote{https://github.com/Aritra23/DragonVerseQA}. The full version of the dataset will be available open-source soon. 

\begin{table}[!ht]
\centering
\caption{\textbf{House of the Dragon} QA pairs by each episode}
\label{table:qadataset1}
\begin{threeparttable}
\resizebox{\columnwidth}{!}{%
\begin{tabular}{|c|c|c|c|m{2.0cm}|}
\hline
\textbf{Ep.} & \textbf{QA Pairs} & \textbf{A. Q. Len. (words)} & \textbf{A. Len. (words)} & \textbf{Notable Characters} \\ \hline
1 & 300 & 15 & 38 & K. Viserys, Rhaenyra, Daemon, Aemma \\ \hline
2 & 300 & 14 & 42 & Daemon, Rhaenyra, Viserys, Otto \\ \hline
3 & 300 & 16 & 39 & Viserys, Rhaenyra, Alicent, Criston \\ \hline
4 & 300 & 16 & 37 & Daemon, Rhaenyra, Viserys, Criston \\ \hline
5 & 300 & 14 & 41 & Criston, Rhaenyra, Daemon, Alicent \\ \hline
6 & 100 & 13 & 36 & Rhaenyra, Criston, Laenor, Viserys \\ \hline
7 & 100 & 15 & 40 & Laena, Corlys, Rhaenys, Rhaenyra, Aemond \\ \hline
8 & 100 & 17 & 38 & Aemond, Corlys, Rhaenys, Viserys, Rhaenyra \\ \hline
9 & 100 & 14 & 43 & Alicent, Otto, Aegon, Criston \\ \hline
10 & 100 & 16 & 39 & Rhaenyra, Aegon, Lucerys, Aemond \\ \hline
\end{tabular}
}
\begin{tablenotes}
      \small
      \item Ep. - Episode, 
      \item A. Q. Len - Avg. Question Length in words
      \item A. Len. - Avg. Answer Length
\end{tablenotes}

\end{threeparttable}
\end{table}

\begin{table}[!ht]
\centering
\caption{\textbf{Game of Thrones} QA pairs by each episode}
\begin{threeparttable}
\resizebox{\columnwidth}{!}{%
\begin{tabular}{|c|c|c|c|m{2.0cm}|}
\hline
\textbf{Ep.} & \textbf{QA Pairs} & \textbf{A. Q. Len. (words)} & \textbf{A. Len. (words)} & \textbf{Notable Characters} \\ \hline
1 & 200 & 14 & 17 & Jon, Daenerys, Sansa, Arya, Tyrion \\ \hline
2 & 200 & 15 & 18 & Jaime, Brienne, Jon, Daenerys, Sansa, Arya, Tyrion \\ \hline
3 & 200 & 15 & 19 & Arya, Jon, Daenerys, Bran, Night King \\ \hline
4 & 200 & 16 & 20 & Daenerys, Jon, Sansa, Tyrion, Cersei, Euron \\ \hline
5 & 200 & 14 & 21 & Daenerys, Jon, Tyrion, Cersei, Jaime, Arya \\ \hline
6 & 200 & 12 & 22 & Jon, Daenerys, Tyrion, Bran, Sansa, Arya \\ \hline

\end{tabular}
}
\label{table:qadataset2}
\begin{tablenotes}
      \small
      \item Ep. - Episode, 
      \item A.Q.Len - Avg. Question Length in words, 
      \item A. Len. - Avg. Answer Length
\end{tablenotes}
\end{threeparttable}
\end{table}

\begin{table}[ht!]
\centering
\caption{DragonVerseQA dataset summary statistics}
\label{tab:summary_metrics}
\resizebox{\columnwidth}{!}{%
\begin{tabular}{|c|m{2.0cm}|}
\hline
\textbf{Metrics} & \textbf{Value} \\ \hline
\textbf{Total Number of QA Pairs} & 3200 \\ \hline
\textbf{Avg. Length of Questions (words)} & 15 \\ \hline
\textbf{Avg. Length of Answers (words)} & 31 \\ \hline
\textbf{Most Frequently Mentioned Characters} & Rhaenyra, Viserys, Daemon, Jon. Daenerys, Arya \\ \hline
\textbf{Predominant Themes} & Power/Politics, Family/Lineage, Morality/Choices \\ \hline
\end{tabular}
}
\end{table}

\begin{figure}
\centering
\includegraphics[width=0.45\textwidth]{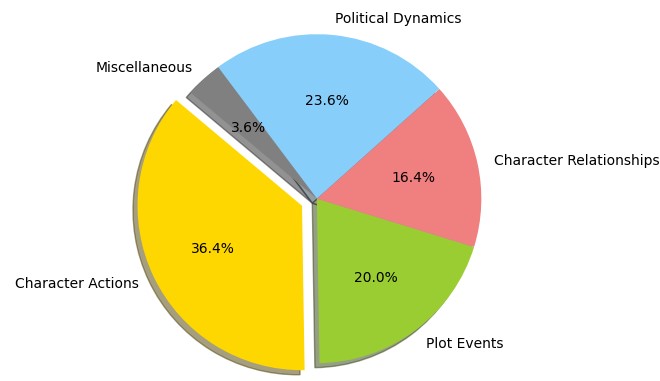}
    \vspace{-2mm}\caption{Factoid Questions Category Distribution}
    \label{figure:factoidqpie}
\end{figure}

We have also analyzed the factoid question types generated and classified them considering common categories such as - a) \textbf{Character actions} (e.g., "What actions does Prince Daemon take in this episode?", b) \textbf{Plot Events} (e.g., "What tragic event occurs during the queen’s labor?"), c) \textbf{Character Relationships} (e.g., "How does the relationship between Rhaenyra and Daemon evolve?"), d) \textbf{Political Dynamics} (e.g., "How does King Viserys handle the issue of his heir?"), e) \textbf{Miscellaneous} ( Questions that don't fit neatly into the above categories). Based on the QA pairs generated, we have demonstrated the distribution as a pie chart in Figure \ref{figure:factoidqpie}. As observed, character actions are the predominant question type.

\subsection{Supportive Knowledge graph}
\begin{figure}
\centering
\includegraphics[width=0.50\textwidth]{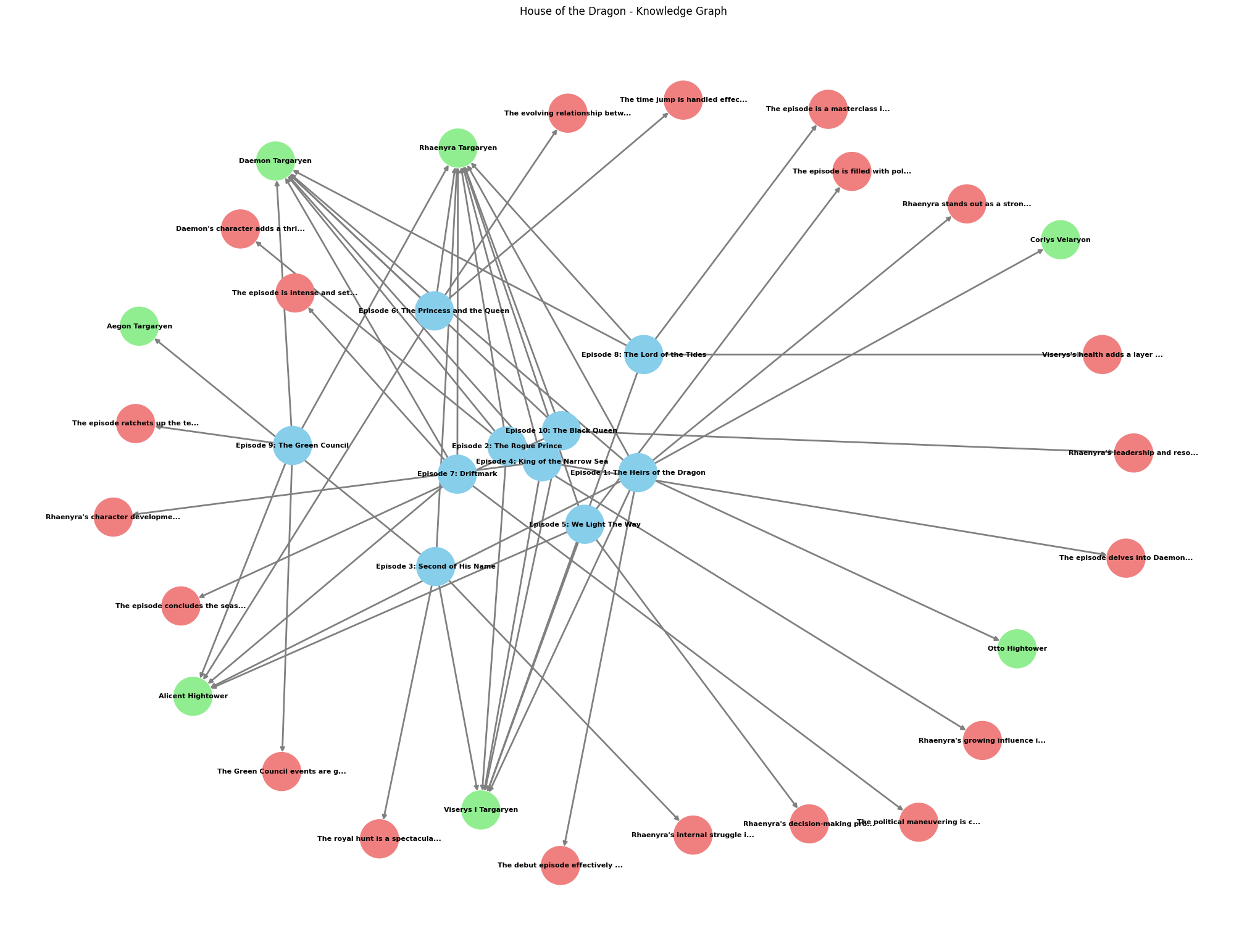}
    \vspace{-2mm}\caption{Knowledge Graph Representation - DragonVerseQA}
    \label{figure:kg}
\end{figure}

We integrate episode narratives, character dynamics, and audience reviews into a knowledge graph to gain deeper insights into interactions, offering a comprehensive view of each episode's impact. This multi-dimensional model, illustrated in Figure \ref{figure:kg}, underscores our analysis's need for richer contextual representation.

Figure \ref{figure:kg} depicts nodes representing episodes, characters, and summarized reviews. Edges connect episodes to characters (labeled as `features`) and episodes to their respective reviews (labeled `reviews`). To enhance clarity, nodes are color-coded by type: episodes in sky blue, characters in light green, and reviews in light coral.

This graph effectively visualizes the intricate narratives and political intrigue inherent to "House of the Dragon" Season 1 episodes, supporting narrative analysis and storytelling research. The graph facilitates sentiment analysis research by linking episodes directly with qualitative audience feedback, ultimately improving recommendation systems and personalization in media content.

\subsection{Comparison with State-of-the-Art Benchmarks}
Given their similarities, we compared the ELI5 (Explain Like I’m 5) long-form question-answering dataset and our proposed DragonVerseQA long-form QA dataset. Table \ref{tab:dataset_comparison_long_form} provides an overview of this comparison. Regarding research applications, ELI5 is well-suited for developing educational tools and AI models that aim to simplify complex topics. Conversely, DragonVerseQA is designed to enhance conversational AI, particularly in the context of understanding and engaging with TV series narratives. Additionally, it holds potential for sentiment analysis based on audience reviews, advanced summarization techniques, and relationship extraction.

\begin{table}[h]
\centering
\caption{Comparison Between ELI5 and DragonVerseQA}
\label{tab:dataset_comparison_long_form}
\begin{adjustbox}{max width=\columnwidth}
\begin{tabular}{|c|c|c|}
\hline
\textbf{Feature / Metric} & \textbf{ELI5 } & \textbf{DragonVerseQA} \\ \hline
\textbf{Source Material} & Reddit & HBO, IMDb, WikiData \\ \hline
\textbf{Context Complexity} & Moderate & High\\ \hline
\textbf{Answer Length (avg)} & Variable, often extensive &  around 40 words\\ \hline
\textbf{Application Domain} & General knowledge & TV series\\ \hline
\textbf{Data Quality} & Varies, user-generated & high, filtered for quality \\ \hline
\textbf{Primary Use Cases} & General QA & Conversational AI \\ \hline
\textbf{Review Integration} & No & Yes, filtered for quality \\ \hline
\end{tabular}
\end{adjustbox}
\end{table}

We also showcased another comparative analysis with benchmark QA datasets like NaturalQuesions(NQ), TriviaQA Long, and SQuAD 2.0 in Table \ref{tab:dataset_comparison_transposed} to highlight our dataset's potential utilities and applications. We compare the datasets across crucial metrics like question diversity, answer length, dataset size, and type of source material.

\begin{table}[h!]
\centering
\caption{Comparison with State-Of-The-Art Datasets}
\label{tab:dataset_comparison_transposed}
\begin{adjustbox}{max width=\columnwidth}
\begin{tabular}{|c|c|m{2.5cm}|}
\hline
\textbf{Dataset} & \textbf{Metric} & \textbf{Value} \\ \hline
\multirow{5}{*}{\textbf{DragonVerseQA}} 
    & \# of QA Pairs & 3200 \\ \cline{2-3} 
    & Answer Length (avg) & 31 words \\ \cline{2-3} 
    & Question Diversity & High: Long-form \& context-specific \\ \cline{2-3} 
    & Source Material & Episode summaries, reviews \\ \cline{2-3} 
    & QA Type & Open-domain, long-form \\ \hline
\multirow{5}{*}{\textbf{NaturalQuestions(NQ)}} 
    & \# of QA Pairs & 307,373 \\ \cline{2-3} 
    & Answer Length (avg) & 22 words \\ \cline{2-3} 
    & Question Diversity & Moderate \\ \cline{2-3} 
    & Source Material & Wikipedia excerpts \\ \cline{2-3} 
    & QA Type & Open-domain \\ \hline
\multirow{5}{*}{\textbf{TriviaQA Long}} 
    & \# of QA Pairs & 650,000 \\ \cline{2-3} 
    & Answer Length (avg) & 35 words \\ \cline{2-3} 
    & Question Diversity & Moderate \\ \cline{2-3} 
    & Source Material & Web documents \\ \cline{2-3} 
    & QA Type & Open-domain \\ \hline
\multirow{5}{*}{\textbf{SQuAD 2.0}} 
    & \# of QA Pairs & 150,000 \\ \cline{2-3} 
    & Answer Length (avg) & 19 words \\ \cline{2-3} 
    & Question Diversity & High \\ \cline{2-3} 
    & Source Material & Wikipedia excerpts \\ \cline{2-3} 
    & QA Type & Context-specific \\ \hline
\end{tabular}
\end{adjustbox}
\end{table}

While smaller than state-of-the-art counterparts, our dataset offers high relevance within the specialized domain of the OTT TV series fantasy universe. Its longer average answer length supports deeper contextual coverage, which is ideal for long-form QA tasks. With high question diversity arising from various contexts across summaries and reviews, it surpasses some leading datasets. Unique source materials, such as episode reviews and summaries, enrich the context beyond what Wikipedia-only datasets can offer.

\section{Evaluation}

\subsection{Key Evaluation Metrics}
We evaluated our dataset against other benchmark datasets using up-to-date evaluation metrics, including F1-score, BLEU (Bilingual Evaluation Understudy Score) \cite{b22}, and ROUGE (Recall-Oriented Understudy for Gisting Evaluation) \cite{b23}, through the \textit{nlgeval}\cite{b24} library . The F1-score assesses the QA pairs' accuracy by balancing precision and recall. BLEU examines how closely the LLM-generated answers align with reference answers, while ROUGE checks the overlap between answers and references. We also introduced a Context Relevance metric, a human-annotated score reflecting how relevant the QA pair is to the source text. This was achieved using two annotators via Amazon Mechanical Turk. Table \ref{tab:qualitative_comparison} presents the scores: a higher F1-score indicates more accurate answer extraction, a higher BLEU score implies a closer match to reference answers, the ROUGE-L score reflects coverage of key points from references, and Context Relevance is rated from 1 (poor relevance) to 5 (excellent relevance).

\begin{table}[h!]
\centering
\caption{Performance Comparison Across Datasets}
\label{tab:qualitative_comparison}
\resizebox{\columnwidth}{!}{%
\begin{tabular}{|c|c|c|c|c|}
\hline
\textbf{Dataset} & \textbf{F1-Score} & \textbf{BLEU} & \textbf{ROUGE-L} & \textbf{Context Relevance} \\ \hline
DragonVerseQA & 85.6 & 0.45 & 0.58 & 4.8 \\ \hline
Natural Questions (NQ) & 78.9 & 0.42 & 0.54 & 4.5 \\ \hline
TriviaQA & 74.3 & 0.40 & 0.50 & 4.2 \\ \hline
SQuAD 2.0 & 82.6 & 0.46 & 0.57 & 4.6 \\ \hline
\end{tabular}
}
\end{table}

\subsection{Dataset Validation: Quality Assurance}
To ensure the quality and comprehensiveness of our generated QA pairs, we adopted a two-fold approach:

1. \textit{Manual Review:}  A subset of annotations was assessed by human experts to evaluate relevance and accuracy.
2. \textit{Automated Validation:} Scripts were used to verify formatting consistency and logical coherence.

For manual review, we randomly selected 200 QA pairs from our dataset and reviewed them by two experts familiar with "House of the Dragon" and "Game of Thrones" TV series. They evaluated each pair on: i) \textit{Relevance}: Is the question relevant to the provided text? , ii) \textit{Correctness}: Is the answer accurately and comprehensively addressing the question? , iii) \textit{Clarity}: Is the question clearly formulated?
Their consolidated assessments are reflected in Table \ref{tab:metrics_evaluation}. High ratings suggest the model could generate meaningful questions and correct answers related to the text. A slightly lower clarity score indicates room for improving question generation capability.

\begin{table}[h!]
\small
\centering
\caption{Manual Evaluation Assessment}
\label{tab:metrics_evaluation}
\resizebox{\columnwidth}{!}{%
\begin{tabular}{|c|c|c|c|c|}
\hline
\textbf{Evaluation Aspect} & \textbf{Very Good (\%)} & \textbf{Good (\%)} & \textbf{Average (\%)} & \textbf{Poor (\%)} \\ \hline
Relevance & 55 & 30 & 10 & 5 \\ \hline
Correctness & 60 & 25 & 10 & 5 \\ \hline
Clarity & 50 & 35 & 10 & 5 \\ \hline
\end{tabular}
}
\end{table}

The automated validation process utilizes scripts to verify format consistency and the suitability of answer lengths. For format consistency, the script ensures that the structure of QA pairs aligns with our predefined schema as outlined in Figure \ref{figure: OTTQA Structure}. In assessing answer length appropriateness, the script identifies either excessively short or long answers that differ from expected standards. The compiled results of these checks are presented in Table \ref{tab:validation_metrics}.

\begin{table}[h!]
\centering
\caption{Automated Validation Assessment}
\label{tab:validation_metrics}
\begin{adjustbox}{max width=\textwidth}
\begin{tabular}{|l|c|c|}
\hline
\textbf{Metric} & \textbf{Pass (\%)} & \textbf{Fail (\%)} \\ \hline
Format Consistency & 98 & 2 \\ \hline
Answer Length & 97 & 3 \\ \hline
\end{tabular}
\end{adjustbox}
\end{table}

\subsection{Ablation Studies}
We employ our Answer-Aware Question Generation Model for our question generation process, which is fine-tuned using BERT-HLSQG \cite{b2}. In our ablation study, this model serves as the baseline, using the BERT encoder and Hierarchical Decoder as core components. We utilize the PyTorch implementation of BERT for the initial setup. Hyperparameters for our experiments include a batch size 32 for both training and validation datasets, with gradient accumulation steps set to 8 and training conducted over 10 epochs. The Adam Optimizer is applied with an initial learning rate of 1e-4. We modify encoder configurations, contextual information, the hierarchical decoder, and various hyperparameters during the ablation experiments to assess their effects.

As shown in Table \ref{tab:encoder_performance}, the Full BERT encoder achieves the baseline performance. The BERT-base exhibits a marginal decrease in performance, likely due to size constraints, while DistilBERT\cite{b32} shows a more pronounced decline, revealing its limited capacity for context comprehension.

\begin{table}[h]
\centering
\caption{Performance with Different BERT encoders}
\label{tab:encoder_performance}
\resizebox{\columnwidth}{!}{%
\begin{tabular}{|c|c|c|c|}
\hline
\textbf{Configuration} & \textbf{BLEU} & \textbf{ROUGE} & \textbf{Exact Match} \\ \hline
Full BERT (Baseline)    &85.2     &88.1      &78.4   \\ \hline
BERT-Base &82.5 & 85.4 & 75.6 \\ \hline
DistilBERT & 78.1 & 80.2 & 71.3 \\ \hline
\end{tabular}
}
\end{table}

Table \ref{tab:contextual-ablation} presents the results of the contextual ablation study, highlighting a decrease in performance when only episode summaries are utilized. This drop occurs because event descriptions in summaries are more detailed, which is crucial for QA pair generation, as opposed to relying solely on user reviews or character lists.

\begin{table}[h!]
\centering
\caption{Performance with Different Contextual Information}
\label{tab:contextual-ablation}
\resizebox{\columnwidth}{!}{%
\begin{tabular}{|c|c|c|c|}
\hline
\textbf{Configuration} & \textbf{BLEU} & \textbf{ROUGE} & \textbf{Exact Match} \\ \hline
Episode Summaries Only (Baseline)   &85.2     &88.1      &78.4   \\ \hline
Character Lists Only & 65.3 & 68.7 & 59.4 \\ \hline
User Reviews Only & 62.7 & 64.8 & 57.1 \\ \hline
\end{tabular}
}
\end{table}

Table \ref{tab:Hierarchical-ablation} illustrates that the hierarchical decoder consistently surpasses the flat decoder across various metrics. This performance indicates its superior ability to manage context and produce coherent outputs.

\begin{table}[h!]
\centering
\caption{Performance with Different Hierarchical Decoder}
\label{tab:Hierarchical-ablation}
\resizebox{\columnwidth}{!}{%
\begin{tabular}{|c|c|c|c|}
\hline
\textbf{Configuration} & \textbf{BLEU} & \textbf{ROUGE} & \textbf{Exact Match} \\ \hline
Standard Hierarchical Decoder (Baseline) & 85.2 & 88.1 & 78.4 \\ \hline
Flat Decoder & 73.9 & 76.5 & 69.2 \\ \hline
No Hierarchical Structuring & 70.4 & 72.3 & 65.8 \\ \hline
\end{tabular}
}
\end{table}

Table \ref{tab:hyperparameter-ablation} details the various hyperparameter configurations tested in our ablation study, along with the corresponding results. This table provides insights into how different settings influence model performance.

\begin{table}[h!]
\centering
\caption{Performance with Different hyperparameter Settings}
\label{tab:hyperparameter-ablation}
\resizebox{\columnwidth}{!}{%
\begin{tabular}{|c|c|c|c|}
\hline
\textbf{Setting} & \textbf{BLEU} & \textbf{ROUGE} & \textbf{Exact Match} \\ \hline
Learning Rate: 1e-5 & 0.50 & 0.55 & 0.65 \\ \hline
Learning Rate: 1e-4 & 0.55 & 0.60 & 0.70 \\ \hline
Batch Size: 16 & 0.48 & 0.52 & 0.64 \\ \hline
Batch Size: 32 & 0.55 & 0.60 & 0.70 \\ \hline
\end{tabular}
}
\end{table}

These findings from the ablation study underscore the diverse factors that enhance the effectiveness of our question generation model, which is fine-tuned using BERT-HLSQG in developing the DragonVerseQA dataset.

\section{Limitation, Conclusion and Future Works}

The development of the DragonVerseQA dataset marks a notable advancement in narrative-focused question-answering systems for television series. Concentrating on the "House of the Dragon" and "Game of Thrones" series, this dataset offers in-depth question-answer pairs exploring character development, plot events, and thematic elements. Its structured approach facilitates the investigation of complex narrative questions and aids in the training and assessment of sophisticated natural language processing models. This dataset enhances our understanding of the series' intricate storytelling and contributes to broader research in long-form, domain-specific question answering.

Despite its benefits, the DragonVerseQA dataset faces certain scope limitations. The dataset caters to the fantasy universe TV series genre and may not apply to other series of a different genre. This dataset version includes a limited number of QA pairs per episode, which might restrict the training of highly generalizable models. Also, initial bias filtering and spam detection rely heavily on the quality of labeled data and pre-trained model effectiveness and can potentially miss subtly biased content.

DragonVerseQA dataset can be enhanced in several future directions:
\begin{itemize}
    \item \textit{Expansion to other genres} - The dataset can be broadened to include other TV series genres to facilitate cross-domain QA model training and evaluation.
    \item \textit{Data-Augmentation} - Dataset volume can be increased by adding more episodes, detailed scene descriptions, and user-generated content like reviews and forum discussions.
    \item \textit{Improved Annotation} -  Invent a more robust annotation process, possibly integrating crowd-sourcing with expert reviews for greater accuracy and consistency.
    \item \textit{Advanced Filtering with Multimodal Data} -  Employment of sophisticated methods for bias detection and spam filtering, utilizing advanced transformers and multi-modal approaches that combine text, audio, and visual data.
    \item \textit{Evaluation metrics} -  Introduction of new evaluation metrics tailored specifically for narrative-driven QA tasks, ensuring that model performance is assessed accurately in the context of complex storytelling.
\end{itemize}

\section*{Acknowledgment}

This research is supported by a grant from the Natural Sciences \& Engineering Research Council (NSERC) of Canada.



\begin{thebibliography}{00}
\bibitem{b1} Devlin, J., Chang, M.W., Lee, K., \& Toutanova, K. (2019). BERT: Pre-training of Deep Bidirectional Transformers for Language Understanding. In Proceedings of the 2019 Conference of the North American Chapter of the Association for Computational Linguistics: Human Language Technologies (NAACL-HLT).
\bibitem{b2} Ying-Hong Chan, Yao-Chung Fan. A Recurrent BERT-based Model for Question Generation, Proceedings of the Second Workshop on Machine Reading for Question Answering, pages 154–162 Hong Kong, China, November 4, 2019.(ACL)
\bibitem{b3} Fan, Angela, Yacine Jernite, Ethan Perez, David Grangier, Jason Weston, and Michael Auli. "ELI5: Long-form question answering." arXiv preprint arXiv:1907.09190 (2019).
\bibitem{b4} Pranav Rajpurkar, Jian Zhang, Konstantin Lopyrev, and Percy Liang. 2016. Squad: 100,000+ questions for machine comprehension of text. In Proceedings of the 2016 Conference on Empirical Methods in Natural Language Processing (EMNLP). ACL, Austin, Texas, pages 2383–2392. 
\bibitem{b5} Joshi, Mandar, Eunsol Choi, Daniel S. Weld, and Luke Zettlemoyer. "Triviaqa: A large scale distantly supervised challenge dataset for reading comprehension." arXiv preprint arXiv:1705.03551 (2017).
\bibitem{b6} Kwiatkowski, Tom, Jennimaria Palomaki, Olivia Redfield, Michael Collins, Ankur Parikh, Chris Alberti, Danielle Epstein et al. "Natural questions: a benchmark for question answering research." Transactions of the Association for Computational Linguistics 7 (2019): 453-466.
\bibitem{b7} Tapaswi, Makarand, Yukun Zhu, Rainer Stiefelhagen, Antonio Torralba, Raquel Urtasun, and Sanja Fidler. "Movieqa: Understanding stories in movies through question-answering." In Proceedings of the IEEE conference on computer vision and pattern recognition, pp. 4631-4640. 2016.
\bibitem{b8} Lei, Jie, Licheng Yu, Mohit Bansal, and Tamara L. Berg. "Tvqa: Localized, compositional video question answering." arXiv preprint arXiv:1809.01696 (2018).
\bibitem{b9} Gomez-Uribe, Carlos A., and Neil Hunt. "The netflix recommender system: Algorithms, business value, and innovation." ACM Transactions on Management Information Systems (TMIS) 6, no. 4 (2015): 1-19.
\bibitem{b10} Hebbar, Rajat, Krishna Somandepalli, and Shrikanth Narayanan. "Robust speech activity detection in movie audio: Data resources and experimental evaluation." In ICASSP 2019-2019 IEEE International Conference on Acoustics, Speech and Signal Processing (ICASSP), pp. 4105-4109. IEEE, 2019.
\bibitem{b11} Raji, Inioluwa Deborah, Emily M. Bender, Amandalynne Paullada, Emily Denton, and Alex Hanna. "AI and the everything in the whole wide world benchmark." arXiv preprint arXiv:2111.15366 (2021).
\bibitem{b12} Kočiský, Tomáš, Jonathan Schwarz, Phil Blunsom, Chris Dyer, Karl Moritz Hermann, Gábor Melis, and Edward Grefenstette. "The narrativeqa reading comprehension challenge." Transactions of the Association for Computational Linguistics 6 (2018): 317-328.
\bibitem{b13} Nguyen, Tri, Mir Rosenberg, Xia Song, Jianfeng Gao, Saurabh Tiwary, Rangan Majumder, and Li Deng. "Ms marco: A human-generated machine reading comprehension dataset." (2016).
\bibitem{b14} Yang, Zhilin, Peng Qi, Saizheng Zhang, Yoshua Bengio, William W. Cohen, Ruslan Salakhutdinov, and Christopher D. Manning. "HotpotQA: A dataset for diverse, explainable multi-hop question answering." arXiv preprint arXiv:1809.09600 (2018).
\bibitem{b15} Choi, Eunsol, He He, Mohit Iyyer, Mark Yatskar, Wen-tau Yih, Yejin Choi, Percy Liang, and Luke Zettlemoyer. "QuAC: Question answering in context." arXiv preprint arXiv:1808.07036 (2018).
\bibitem{b16} Reddy, Siva, Danqi Chen, and Christopher D. Manning. "Coqa: A conversational question answering challenge." Transactions of the Association for Computational Linguistics 7 (2019): 249-266.
\bibitem{b17}Yang, Yi, Wen-tau Yih, and Christopher Meek. "Wikiqa: A challenge dataset for open-domain question answering." In Proceedings of the 2015 Conference on Empirical Methods in Natural Language Processing, pp. 2013-2018. 2015.
\bibitem{b18} https://beautiful-soup-4.readthedocs.io/en/latest/
\bibitem{b19} https://huggingface.co/mrm8488/bert-tiny-finetuned-sms-spam-detection
\bibitem{b20} https://textblob.readthedocs.io/en/dev/
\bibitem{b21}Seonwoo, Yeon, Ji-Hoon Kim, Jung-Woo Ha, and Alice Oh. "Context-aware answer extraction in question answering." arXiv preprint arXiv:2011.02687 (2020).
\bibitem{b22}Kishore Papineni, Salim Roukos, Todd Ward, and Wei-Jing Zhu. 2002. Bleu: a method for automatic evaluation of machine translation. In Proceedings of 40th Annual Meeting of the Association for Computational Linguistics. Association for Computational Linguistics, Philadelphia, Pennsylvania, USA, pages 311–318. https://doi.org/10.3115/1073083.1073135.
\bibitem{b23}Lin, Chin-Yew. "Rouge: A package for automatic evaluation of summaries." Text summarization branches out. 2004.
\bibitem{b24}Indurthi, Sathish Reddy, et al. "Generating natural language question-answer pairs from a knowledge graph using an based question generation model." Proceedings of the 15th Conference of the European Chapter of the Association for Computational Linguistics: Volume 1, Long Papers. 2017.
\bibitem{b25}Hui, Z., Liu, X., \& Sun, M. (2017, December). A Context-aware Attention Network for Interactive Question Answering. In Proceedings of the 55th Annual Meeting of the Association for Computational Linguistics (Volume 1: Long Papers) (pp. 1212-1222).
\bibitem{b26}Cambazoglu, B. Barla, et al. "A Review of Public Datasets in Question Answering Research."
\bibitem{b27}Raffel, Colin, et al. "Exploring the limits of transfer learning with a unified text-to-text transformer." arXiv preprint arXiv:1910.10683 (2019).
\bibitem{b28}Xiong, Wenhan, et al. "TWEETQA: A social media focused question answering dataset." arXiv preprint arXiv:1907.06292 (2019).
\bibitem{b29} Lahiri, Aritra Kumar, and Qinmin Vivian Hu. "Gameofthronesqa: Answer-aware question-answer pairs for tv series." In European Conference on Information Retrieval, pp. 180-189. Cham: Springer International Publishing, 2022.
\bibitem{b30} Chen, Jiawei, Hongyu Lin, Xianpei Han, and Le Sun. "Benchmarking large language models in retrieval-augmented generation." In Proceedings of the AAAI Conference on Artificial Intelligence, vol. 38, no. 16, pp. 17754-17762. 2024.
\bibitem{b31}Xinya Du and Claire Cardie. 2018. Harvesting Paragraph-level Question-Answer Pairs from Wikipedia. In Proceedings of the 56th Annual Meeting of the Association for Computational Linguistics (Volume 1: Long Papers). 1907–1917.
\bibitem{b32}Sanh, V. (2019). DistilBERT, A Distilled Version of BERT: Smaller, Faster, Cheaper, and Lighter. arXiv preprint arXiv:1910.01108.
\bibitem{b33} Radford, A., Wu, J., Child, R., Luan, D., Amodei, D., \& Sutskever, I. (2019). Language Models are Unsupervised Multitask Learners—OpenAI blog.
\bibitem{b34} Chen, Q., Zhu, X., Ling, Z.H., Wei, S., \& Jiang, H. (2019). Context-aware question answering over dialogue with attention-based LSTM networks. In Proceedings of the 2019 Conference on Empirical Methods in Natural Language Processing and the 9th International Joint Conference on Natural Language Processing (EMNLP-IJCNLP).
\bibitem{b35} Brownlee, J., Thorne, J., Houlsby, N., Groot, R., \& Clark, P. (2019). Question Answering with Neural Attention-based Hierarchical Contexts. In Proceedings of the 2019 Conference on Empirical Methods in Natural Language Processing and the 9th International Joint Conference on Natural Language Processing (EMNLP-IJCNLP).
\bibitem{b36}Lopez, Luis Enrico, et al. "Simplifying paragraph-level question generation via transformer language models." PRICAI 2021: Trends in Artificial Intelligence: 18th Pacific Rim International Conference on Artificial Intelligence, PRICAI 2021, Hanoi, Vietnam, November 8–12, 2021.
\bibitem{b37}Tong Wang, Xingdi Yuan, and Adam Trischler. 2017. A joint model for question answering and question generation. arXiv preprint arXiv:1706.01450 (2017).
\bibitem{b38}Nan Duan, Duyu Tang, Peng Chen, and Ming Zhou. 2017. Question generation for question answering. In Proceedings of the 2017 Conference on Empirical Methods in Natural Language Processing, pages 866–874.. http://www.aclweb.org/anthology/D13-1160.
\bibitem{b39}Sandeep Subramanian, Tong Wang, Xingdi Yuan, Saizheng Zhang, Adam Trischler, and Yoshua Bengio. 2018. Neural Models for Key Phrase Extraction and Question Generation. In Proceedings of the Workshop on Machine Reading for Question Answering. 78–88.
\bibitem{b40}Liangming Pan, Wenqiang Lei, Tat-Seng Chua, and Min-Yen Kan. 2019. Recent advances in neural question generation. arXiv preprint arXiv:1905.08949 (2019).
\bibitem{b41}Xingwu Sun, Jing Liu, Yajuan Lyu, Wei He, Yanjun Ma, and Shi Wang. 2018. Answer-focused and position-aware neural question generation. In Proceedings of the 2018 Conference on Empirical Methods in Natural Language Processing. 3930–3939.
\bibitem{b42}Cui, Shaobo, et al. "Onestop qamaker: extract question-answer pairs from text in a one-stop approach." arXiv preprint arXiv:2102.12128 (2021).
\bibitem{b43}Pum-Mo Ryu, Myung-Gil Jang, and Hyun-Ki Kim.2014. Open domain question answering using Wikipedia-based knowledge model. Information Processing and Management, 50(5):683 – 692.
\bibitem{b44}Mou, Xiangyang, Chenghao Yang, Mo Yu, Bingsheng Yao, Xiaoxiao Guo, Saloni Potdar, and Hui Su. "Narrative question answering with cutting-edge open-domain qa techniques: A comprehensive study." Transactions of the Association for Computational Linguistics 9 (2021): 1032-1046.
\bibitem{b45}Chen, Anthony, Gabriel Stanovsky, Sameer Singh, and Matt Gardner. "Evaluating question answering evaluation." In Proceedings of the 2nd workshop on machine reading for question answering, pp. 119-124. 2019.
\bibitem{b46}Choi, Seongho, et al. "Dramaqa: Character-centered video story understanding with hierarchical qa." Proceedings of the AAAI Conference on Artificial Intelligence. Vol. 35. No. 2. 2021.
\bibitem{b47}Lal, Y. K., Chambers, N., Mooney, R., \& Balasubramanian, N. (2021). TellMeWhy: A dataset for answering why-questions in narratives. arXiv preprint arXiv:2106.06132.
\bibitem{b48}Xu, Y., Wang, D., Yu, M., Ritchie, D., Yao, B., Wu, T., ... \& Warschauer, M. (2022). Fantastic Questions and Where to Find Them: FairytaleQA--An Authentic Dataset for Narrative Comprehension. arXiv preprint arXiv:2203.13947.
\bibitem{b49}Izacard, Gautier, and Edouard Grave. "Distilling knowledge from reader to retriever for question answering." arXiv preprint arXiv:2012.04584 (2020).
\bibitem{b50}Nan, Linyong, Chiachun Hsieh, Ziming Mao, Xi Victoria Lin, Neha Verma, Rui Zhang, Wojciech Kryściński et al. "FeTaQA: Free-form table question answering." Transactions of the Association for Computational Linguistics 10 (2022): 35-49.


\end{thebibliography}
\end{document}